\documentclass{article}

\usepackage{iclr2027_conference,times}
\usepackage{wrapfig}
\usepackage[utf8]{inputenc}
\usepackage[T1]{fontenc}
\usepackage{hyperref}
\hypersetup{hidelinks,pdftitle={Decision Transformer with Task-Aware Representations for Offline Meta-Reinforcement Learning},pdfauthor={Xinbo Zhao, Yuhang Liu, Xin Zhang, Lei Le, Zhaohan Xi, Yingxue Zhang}}
\usepackage{url}
\usepackage{booktabs}
\usepackage{amsfonts}
\usepackage{amsmath}
\usepackage{amssymb}
\usepackage{mathtools}
\usepackage{amsthm}
\usepackage{nicefrac}
\usepackage{microtype}
\usepackage{xcolor}
\usepackage{graphicx}
\usepackage{subcaption}
\usepackage{multirow}
\usepackage{algorithm}
\usepackage{enumitem}
\usepackage{algorithmic}
\usepackage{placeins}
\usepackage{graphicx}
\usepackage[capitalize,noabbrev]{cleveref}

\theoremstyle{plain}

\theoremstyle{definition}

\theoremstyle{remark}

\title{Decision Transformer with Task-Aware\\
Representations for Offline\\
Meta-Reinforcement Learning}

\author{%
\begin{minipage}[t]{\dimexpr\textwidth-2\tabcolsep\relax}
\centering
\normalfont\normalsize
\textbf{Xinbo Zhao}$^{1}$, \textbf{Yuhang Liu}$^{1}$, \textbf{Xin Zhang}$^{2}$,\\
\textbf{Lei Le}$^{3}$, \textbf{Zhaohan Xi}$^{1}$, \textbf{Yingxue Zhang}$^{1}$\\[0.4em]
\small
$^{1}$State University of New York at Binghamton\\
$^{2}$San Diego State University\\
$^{3}$LinkedIn\\[0.35em]
\footnotesize
\texttt{\{xzhao9,yliu41,zxi1,yzhang42\}@binghamton.edu}\\
\texttt{xzhang19@sdsu.edu}\quad\texttt{some.other.lei@gmail.com}
\end{minipage}}

\iclrfinalcopy

\begin{document}

\maketitle
\lhead{Preprint}

\begin{abstract}
To enable effective generalization in offline meta-RL, it is essential to accurately infer the underlying task through latent task representations, which provide a compact and transferable summary of task-specific characteristics for guiding policy generation. 
In this paper, we propose \textbf{Contrastive Task-Aware Decision Transformer (CoTA-DT)}, a DT-based offline meta-RL framework that learns task representations by jointly integrating four complementary components: \textit{(i)} contrastive learning from offline trajectories, \textit{(ii)} alignment with natural-language task descriptions, \textit{(iii)} one-step supervision via a task-conditioned world model, and \textit{(iv)} long-term supervision via a task-conditioned Q function. 
The learned task representations are used as conditioning signals for a DT-based policy, enabling task-aware meta-policy learning. To further support generalization in both zero-shot and few-shot settings, we design a language-grounded, self-adaptive strategy that enables effective and reliable adaptation to unseen tasks. 
Notably, in CoTA-DT, the pretrained LLM serves as a shared backbone for language encoding, world modeling, Q-function estimation, and policy learning. This unified design facilitates parameter sharing, improving both efficiency and generalization. 
Extensive experiments across diverse offline meta-RL benchmarks demonstrate that CoTA-DT consistently outperforms prior methods, particularly in challenging zero-shot generalization scenarios.
\end{abstract}

\section{Introduction}
\label{sec:intro}

Offline meta-reinforcement learning (RL) aims to learn transferable priors from offline trajectories across related tasks, enabling efficient adaptation to new tasks with little or no additional data. In recent years, a prominent line of work in offline meta-RL has built on Decision Transformers (DTs)~\citep{chen2021decision}, leveraging the strong generalization capabilities of transformer architectures~\citep{mitchell2021offline,lin2022context}. These DT-based approaches formulate policy learning as sequence modeling over trajectories and perform adaptation through sequence conditioning.

For example, Prompt-DT~\citep{xu2022prompting} leverages expert demonstrations as prompts to encode task-specific information, enabling prompt-based adaptation at test time. Generalized Decision Transformers (GDT)~\citep{furuta2021,wang2023t3gdt} incorporate hindsight signals, such as reward statistics or retrospective summaries, as conditioning inputs for task adaptation. Meta-DT~\citep{wang2024meta} conditions the policy on compact task representations learned from trajectories to support generalization. 
More recently, T2DA~\citep{t2da2025} conditions a DT-based policy on text embeddings derived from natural-language task descriptions.
Despite the advances, these approaches typically learn task identifiers from a single source—such as language, trajectory data, or reward signals—and use them as conditioning inputs to guide DT-based policies for generalization. When such signals are ambiguous, noisy, or distributionally shifted relative to training, task identification can become unreliable, leading to degraded adaptation performance. These limitations naturally raise a question: \emph{How can we learn an offline meta-policy that achieves reliable task identification and supports robust generalization in both few-shot and zero-shot settings?}

Reliable task identification across a broad family of related tasks depends on the availability of task-specific information. In offline meta-RL, collected trajectories serve as the primary source of information and implicitly encode task differences. However, relying solely on offline data is often insufficient, especially when the data are limited, noisy, or provide only partial coverage of the task space.
Beyond trajectories, additional sources of information are often available. For example, natural-language task descriptions derived from task-level metadata—such as task goals, names, relevant objects, and environment constraints—provide complementary high-level semantics that are not always recoverable from trajectories alone. Meanwhile, environment transitions and rewards vary across tasks, offering informative one-step signals of task-specific dynamics and objectives, while cumulative returns (\emph{e.g.},  Q functions) capture long-horizon goals and provide global signals for distinguishing tasks.
These observations motivate jointly leveraging multiple sources of information to learn robust task representations and enable reliable task identification in offline meta-RL.

In this paper, we propose \textbf{Contrastive Task-Aware Decision Transformer (CoTA-DT)}, a DT-based offline meta-RL framework that learns task representations by jointly integrating four complementary components: \textit{(i)} contrastive task encoding from offline trajectories, \textit{(ii)} alignment with natural-language task descriptions, \textit{(iii)} one-step supervision via a task-conditioned world model, and \textit{(iv)} long-term supervision via a task-conditioned Q function.
Specifically, task representations are first learned from trajectories via a contrastive objective that pulls together representations from the same task while separating those from different tasks. However, trajectory-only representations can be noisy due to data variability. To better capture underlying task objectives and reduce bias, we further align them with natural-language task descriptions encoded by a pretrained large language model (LLM), promoting semantic consistency between trajectory- and language-derived information.
To further refine the representations, we introduce a task-conditioned world model that provides one-step supervision through transitions and rewards, along with a task-conditioned Q function that offers long-horizon supervision over task objectives. These complementary signals regularize the representations, and all components are optimized jointly to learn task representations from multiple perspectives.

The resulting task representations are used as conditioning signals for a DT-based policy, enabling task-aware meta-policy learning. To support generalization in both zero-shot and few-shot settings, we design a language-grounded, self-adaptive strategy for reliable adaptation.
Notably, in CoTA-DT, the pretrained LLM serves as a shared backbone for language encoding, world modeling, Q-function estimation, and policy learning. This unified design enables parameter sharing across stages and leverages the LLM’s prior knowledge from large-scale pretraining, along with its strong sequence modeling and transfer capabilities under prompt conditioning~\citep{brown2020gpt3,ouyang2022training,touvron2023llama}, thereby improving both efficiency and generalization.
\emph{Our main contributions are as follows:}
\begin{itemize}[nosep, leftmargin=*]
    \item We introduce CoTA-DT, an offline meta-RL framework whose policy conditions on a task latent that supports both language-based initialization and updates from observed transitions.

    \item We learn discriminative, language-grounded task representations
that capture task dynamics and long-horizon returns, supporting
adaptation to unseen tasks with a shared LoRA-adapted LLM backbone.

    \item We develop zero-shot deployment through language-grounded latent inversion and few-shot adaptation through support-based initialization, with optional policy fine-tuning.

    \item Extensive experiments across various offline meta-RL benchmarks demonstrate that CoTA-DT consistently outperforms prior methods, particularly in challenging zero-shot generalization settings\footnote{Code available at \url{https://anonymous.4open.science/r/CotaDT-35DB}.}.
\end{itemize}

\section{Related Work}
\label{sec:related}

\noindent\textbf{Offline Meta-RL.}
Offline RL learns policies from fixed datasets without further environment interaction~\citep{levine2020offline,SemiRewardOfflineRL}. Extending this setting, offline meta-RL aims to learn transferable priors that enable generalization across tasks using only offline data~\citep{mitchell2021offline,OfflineMeta2,OfflineMeta3,OfflineMeta4}. 
A large body of work in this area builds on temporal-difference (TD) learning, often incorporating optimization-based meta-learning or latent context inference mechanisms to capture task-specific information~\citep{MAML,mitchell2021offline,xu2018meta,rakelly2019efficient,zintgraf2021varibad,yuan2022robust,li2024towards}. However, applying TD learning in the offline setting is challenging due to the deadly triad~\citep{levine2020offline}, which can lead to instability and error accumulation. To mitigate this issue, many approaches impose constraints to keep learned policies within the support of the dataset~\citep{ajay2022conditional,CQL,IQL,MOPO,MOReL}. 
Other works pursue more stable alternatives, such as return-conditioned or sequence modeling approaches based on Decision Transformers~\citep{wang2024meta,hu2024harmodt,xu2022prompting,wang2023t3gdt,t2da2025}.

\paragraph{LLMs for RL.}
Recent work explores leveraging the representation and generalization capabilities of LLMs for RL, including using them as policy backbones or for reward modeling~\citep{morad2024languageconditionedofflinerlmultirobot,reed2022generalist,brohan2023rt}. 
In the offline RL setting, LLMs are incorporated in several ways. LaMo~\citep{shi2023unleashing} demonstrates that pretrained language models can be adapted as effective policy backbones, while KALM~\citep{pang2404knowledgeable} leverages LLM-generated rollouts to augment offline datasets. More recently, T2DA~\citep{t2da2025} conditions decision models on text embeddings derived from LLMs to support generalization in offline meta-RL, and LLMDPD~\citep{liu2024llm_diffusion} further enhances generalization by integrating LLM priors into diffusion-based policy learning.
In contrast, our method leverages LLMs as a shared backbone for both task representation learning and policy modeling.

\section{Preliminaries}
\subsection{Offline Meta-Reinforcement Learning}
An RL task is formulated as an MDP $M=(S, A, T, r, \gamma)$ with state space $S$, action space $A$, transition $T: S \times A \rightarrow \Delta(S)$, reward $r: S \times A \rightarrow \mathbb{R}$, and discount $\gamma \in [0,1)$. The objective is to learn a policy $\pi(a|s)$ that maximizes the expected return:
$\max_{\pi} J_M(\pi) = \mathbb{E}_{\pi} \left[ \sum_{t=0}^{\infty} \gamma^t r(s_t, a_t) \right]$.
In offline meta-RL, tasks $M_i = (S, A, T_i, r_i, \gamma) \sim P(\mathcal{M})$ share the same state and action spaces but differ in $T_i$ and $r_i$. For each of $N$ training tasks, an offline dataset $\mathcal{D}_i = \{(s_{i,j}, a_{i,j}, r_{i,j}, s'_{i,j})\}_{j=1}^{J}$ is collected by an arbitrary behavior policy $\pi_{\beta}^i$. Each training task is paired with a description $h_i$. For an unseen task $\mathcal{M}_\star\sim P(\mathcal{M})$, the description $h_\star$ is available. Zero-shot evaluation provides no target-task support trajectories before rollout; few-shot evaluation additionally provides $k$ support trajectories. During rollout, the agent observes states and rewards and may update its task latent from past transitions, with policy parameters fixed. The goal is to learn a meta-policy maximizing 
$J(\pi_{\text{meta}}) = \mathbb{E}_{M \sim P(\mathcal{M})} \left[ J_{M}(\pi_{\text{meta}}) \right]$.

\subsection{Decision Transformer}
Decision Transformer (DT) recasts RL as return-conditioned sequence modeling: each trajectory is a sequence of $(R_t, s_t, a_t)$ triplets where $R_t = \sum_{t'=t}^{T} \gamma^{t'-t} r_{t'}$ is the return-to-go (RTG), and the model is trained to autoregressively predict $a_t$ from past RTGs, states, and actions by minimizing prediction error against ground-truth actions.

\begin{figure*}[t]
    \centering
    \includegraphics[width=1\linewidth]{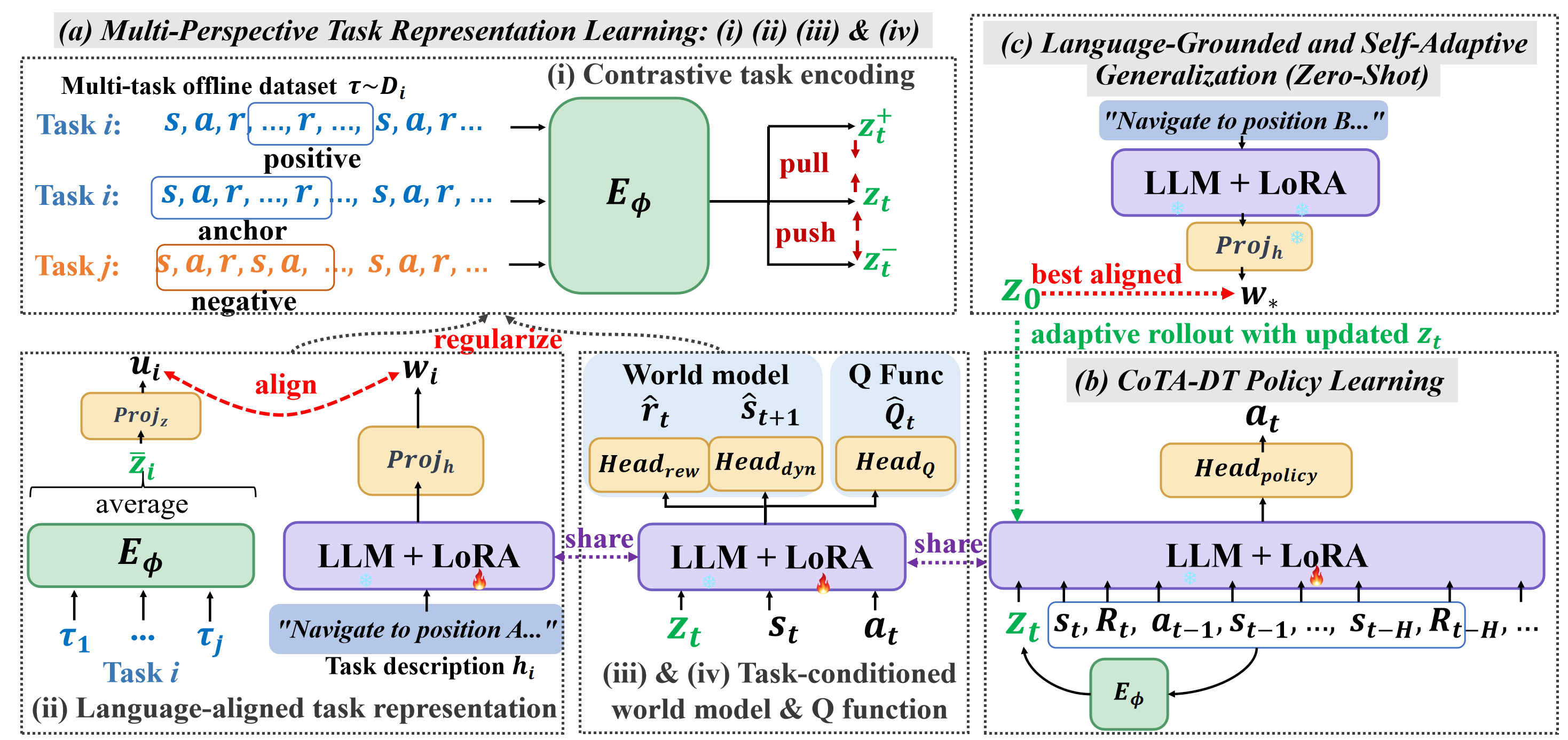}
    \vspace{-2mm}
\caption{\textbf{Overview of CoTA-DT.}
\emph{(a)} Multi-perspective task representation learning jointly optimizes four signals---\emph{(i)} contrastive task encoding from offline trajectories, \emph{(ii)} language alignment between $u_i$ and $w_i$, \emph{(iii)} a task-conditioned world model predicting $\hat r_t, \hat s_{t+1}$, and \emph{(iv)} a long-horizon Q head predicting $\hat Q_t$---all sharing one frozen LLM with a single LoRA adapter.
\emph{(b)} A Decision Transformer policy is trained on the same LLM backbone, conditioned on $z_t$.
\emph{(c)} At zero-shot adaptation, a new task description is encoded into $w_\star$, $z_0$ is constructed by aligning with $w_\star$, and the rollout uses self-adaptive $z_t$ updates.}
    \label{fig:overview}
    \vspace{-4mm}
\end{figure*}

\section{Method}

In this section, we present Contrastive Task-Aware Decision Transformer (CoTA-DT), an offline meta-RL framework that enables reliable task identification through multi-perspective task representations and integrates task inference into DT-based policy for improved generalization.  Fig.~\ref{fig:overview} illustrates the method overview.

\subsection{Multi-Perspective Task Representation Learning}\label{sec:multi-perspective}

To enable effective generalization in offline meta-RL, it is essential to distinguish between tasks by capturing both their differences and shared structure. Accurately inferring the underlying task through latent task representations is therefore critical, as it provides a compact and transferable summary of task-specific characteristics that can be used to condition policy generation. %

However, relying on a single source of information, such as offline datasets, is often insufficient for learning reliable task representations. Offline datasets typically contain trajectories exhibiting varying levels of expertise, data quality, and coverage. As a result, the learned representations can be noisy and less robust.
To address this limitation, we propose to learn and refine latent task representations from multiple perspectives: \textit{(i)} contrastive task encoding from offline trajectories (Sec.~\ref{sec:task-encoding}), \textit{(ii)} alignment with natural-language task descriptions (Sec.~\ref{sec:align}), \textit{(iii)} one-step supervision via a task-conditioned world model (Sec.~\ref{sec:world-model}), and \textit{(iv)} long-term supervision via a task-conditioned Q function (Sec.~\ref{sec:world-model}). %

\subsubsection{Contrastive Task Encoding}\label{sec:task-encoding}
To learn latent task representations, offline trajectories provide a primary source of information for capturing task-specific characteristics, as they inherently reflect the underlying task through observed behaviors. 
We encode short trajectory segments using a separate GRU context encoder $E_\phi$, which maps transitions to a task latent $z\in\mathbb{R}^{64}$; the LLM backbone processes text and projected continuous tokens. Sub-trajectories capture partial yet informative aspects of the task, enabling the learned representations to reflect not only the underlying task objectives but also the stage of task execution.
For example, in locomotion tasks, a short segment can indicate both the objective (\emph{e.g.}, moving forward or turning) and the current execution stage (\emph{e.g.}, initiating motion, stabilizing, or adjusting direction). %

Formally, given a sub-trajectory $\tau_{t-H:t} = \{(s_j, a_j, r_j, s_{j+1})\}_{j=t-H}^{t}$ of horizon $H$ extracted from an offline trajectory, the encoder maps it to a latent representation
$z_t = E_\phi(\tau_{t-H:t}), z_t \in \mathbb{R}^{d_z}$.
For timesteps near the beginning of a trajectory where $t-H < 0$, we construct $\tau_{t-H:t}$ by zero-padding the missing prefix.
For each anchor sub-trajectory $\tau_{t-H:t}$, we construct a positive sample $\tau^{+}$ as a nearby segment from the same trajectory, \emph{e.g.}, $\tau^{+} = \tau_{t-H+\delta:t+\delta}$ with a small offset $\delta$. Negative samples $\{\tau_k^{-}\}_{k=1}^{K}$ are drawn from trajectories of different tasks. Let $z_t^{+} = E_\phi(\tau^{+}), z_{k}^{-} = E_\phi(\tau_k^{-})$.
We optimize the standard InfoNCE~\citep{infoNCE} objective:
\begin{equation}
\mathcal{L}_{\text{NCE}} = -\mathbb{E} \left[ \log \frac{\exp(\mathrm{sim}(z_t,z_t^{+})/\kappa_\alpha)}{\exp(\mathrm{sim}(z_t,z_t^{+})/\kappa_\alpha) + \sum_{k=1}^{K} \exp(\mathrm{sim}(z_t,z_k^{-})/\kappa_\alpha)} \right],
\end{equation}
where $\mathrm{sim}(\cdot,\cdot)$ denotes cosine similarity and $\kappa_\alpha$ is a temperature parameter.
This objective encourages representations of temporally adjacent sub-trajectories from the same task to align, as they share the same task objective and similar execution stages, while pushing apart representations from different tasks. %

\subsubsection{Language-Aligned Task Representation}\label{sec:align}

Since the latent representations $z$ learned by the contrastive encoder are derived from sub-trajectories, they can be biased toward local behaviors or the current execution stage rather than the underlying task objective. This issue is particularly pronounced when sub-trajectories are sampled from low-quality, non-expert data, or from very early stages of task execution, where the task goal is not yet clearly manifested.
To encourage the latent representations to better capture the global task objective, we ground $z$ in natural-language task descriptions, enforcing consistency between latent representations and their corresponding semantic descriptions. 

Specifically, natural-language descriptions $h_i$ are deterministically constructed from task-level metadata of task $M_i$, including the task name, goal, relevant objects, environmental conditions, and typical performance. 
Each description $h_i$ is encoded into a text embedding $v_i$ using a pretrained LLM with LoRA adaptation~\citep{hu2022lora}, \emph{i.e.}, $v_i = \mathrm{LLM}_{\theta_{\mathrm{base}}, \theta_{\mathrm{L}}}\!\big(h_i\big)$, where the LLM backbone $\theta_{\mathrm{base}}$ is kept frozen and the LoRA adapter $\theta_{\mathrm{L}}$ is fine-tuned.  For each task $i$, we also compute an aggregated representation $\bar{z}_i$ by averaging latent representations over a set of sub-trajectories sampled from the offline dataset $\mathcal{D}_i$:
$\bar{z}_i = \frac{1}{m} \sum_{j=1}^{m} E_\phi(\tau^i_j), \tau^i_j \sim \mathcal{D}_i$.
Both $\bar{z}_i$ and $v_i$ are then projected into a shared embedding space via lightweight projection heads $\mathrm{Proj}_z$ and $\mathrm{Proj}_h$ for alignment:
\begin{equation}
u_i = \frac{\mathrm{Proj}_z(\bar{z}_i)}{\|\mathrm{Proj}_z(\bar{z}_i)\|_2}, \qquad 
w_i = \frac{\mathrm{Proj}_h(v_i)}{\|\mathrm{Proj}_h(v_i)\|_2}, \qquad 
u_i, w_i \in \mathbb{R}^{d_a}.
\end{equation}
We then quantify the similarity between $u_i$ and $w_i$ by defining a cosine similarity matrix over a minibatch of tasks $\mathcal{I}$, scaled by a learnable temperature $\kappa_\beta$ optimised in log-space to enforce positivity, following~\citep{clip}:
\begin{equation}
S_{ij} = \frac{u_i^\top w_j}{\kappa_\beta}, \qquad 
\kappa_\beta = e^{\log \kappa_\beta},
\end{equation}

We optimize a symmetric CLIP-style~\citep{clip} alignment objective:
\begin{equation}
\begin{aligned}
\mathcal{L}_{\mathrm{align}} &= \tfrac{1}{2} \left( \mathcal{L}_{z \rightarrow h} + \mathcal{L}_{h \rightarrow z} \right), \\
\mathcal{L}_{z \rightarrow h} &= -\tfrac{1}{|\mathcal{I}|} \sum_{i \in \mathcal{I}} \log \tfrac{\exp(S_{ii})}{\sum_{j \in \mathcal{I}} \exp(S_{ij})}, \quad
\mathcal{L}_{h \rightarrow z} = -\tfrac{1}{|\mathcal{I}|} \sum_{i \in \mathcal{I}} \log \tfrac{\exp(S_{ii})}{\sum_{j \in \mathcal{I}} \exp(S_{ji})}.
\end{aligned}
\label{eq:clip_task_level}
\end{equation}

This contrastive objective treats $(\bar{z}_i, h_i)$ as a positive pair, while $(\bar{z}_i, h_j)$ for $j \neq i$ serve as negatives. It encourages trajectory-derived task representations $z$ to align with their corresponding language descriptions, thereby grounding the latent representations in high-level task semantics.

\subsubsection{Task-Conditioned World Modeling and Q Learning} \label{sec:world-model}
Beyond trajectories and natural-language task descriptions, environment transitions and rewards inherently vary across tasks, providing informative one-step signals of task-specific dynamics and objectives that help regularize task representations. Over longer horizons, cumulative returns (\emph{e.g.}, Q functions) capture task goals and provide complementary global signals for distinguishing tasks.
Motivated by these observations, we introduce a task-conditioned world model and a task-conditioned Q function to further refine the learned representations. 

\paragraph{Task-Conditioned World Modeling.}
Given a transition $(s_t, a_t, r_t, s_{t+1})$ sampled from a sub-trajectory $\tau_{t-H:t}$, we first infer a latent representation $z_t = E_\phi(\tau_{t-H:t})$. 
Conditioned on $z_t$, the world model predicts the immediate reward and next state, \emph{i.e.},
$p\big(r_t, s_{t+1} \mid s_t, a_t;\, z_t\big)$,
and is trained with supervised objectives
\begin{equation}
\mathcal{L}_{\mathrm{rew}} = \mathbb{E}\big[\| \hat r_t - r_t \|_2^2\big], 
\qquad
\mathcal{L}_{\mathrm{dyn}} = \mathbb{E}\big[\| \hat s_{t+1} - s_{t+1} \|_2^2\big],
\end{equation}
where $\hat r_t$ and $\hat s_{t+1}$ denote the predicted reward and next state, respectively.

\paragraph{Task-Conditioned Q Learning.}
To incorporate long-horizon supervision, we introduce a task-conditioned Q function that predicts the return-to-go given a state-action pair $(s_t, a_t)$ and the corresponding task representation $z_t = E_\phi(\tau_{t-H:t})$: 
$Q(s_t, a_t;\, z_t) \approx \mathbb{E}\Big[\sum_{t'=t}^{T} \gamma^{t'-t} r_{t'}\Big]$.
The model is trained with a supervised regression objective:
\begin{equation}
\mathcal{L}_{\mathrm{Q}} = \mathbb{E}\big[\| \hat{R}_t - R_t \|_2^2\big],
\end{equation}
where $R_t$ denotes the empirical return-to-go computed from offline trajectories, and $\hat{R}_t$ is the corresponding prediction produced by the Q function.

Both the world model and the Q function are implemented by repurposing the same pretrained LLM backbone $\theta_{\mathrm{base}}$ with the shared LoRA adapter $\theta_{\mathrm{L}}$, originally fine-tuned for language processing, and augmenting it with task-specific prediction heads for reward, dynamics, and return estimation (shown in Fig.~\ref{fig:overview}).
Moreover, the supervision signals from both components are backpropagated through $z_t$ to jointly update the contrastive task encoder $E_\phi$ and LoRA parameters $\theta_{\mathrm{L}}$. The world model provides one-step supervision through task-specific transitions and rewards, while the Q function supplies long-horizon supervision that captures task objectives over extended trajectories. 
These complementary signals encourage $E_\phi$ to capture both short-term dynamics and long-term goals, improving its discriminative power for task identification and generalization, while also enabling the LLM to better understand environment dynamics and task objectives.

\paragraph{Overall Objective.}
We jointly optimize the multi-perspective representation learning objective:
\begin{equation}
\mathcal{L}_{\mathrm{repr}} = 
\lambda_{1} \mathcal{L}_{\mathrm{NCE}} 
+ \lambda_{2} \mathcal{L}_{\mathrm{align}} 
+ \lambda_{3} \left( \mathcal{L}_{\mathrm{rew}} + \alpha \mathcal{L}_{\mathrm{dyn}} \right) 
+ \lambda_{4} \mathcal{L}_{\mathrm{Q}},
\label{eq:overall_repr}
\end{equation}
where $\lambda_{1}$--$\lambda_{4}$ are balancing coefficients.
This objective enables the task encoder $E_\phi$ to learn reliable, task-aware latent representations $z$ by integrating complementary supervision signals, while the shared LoRA-adapted LLM backbone jointly captures trajectory information, task semantics, environment dynamics, and long-horizon objectives, ultimately benefiting downstream DT-based policy learning.

\subsection{Task-Aware Policy Learning for CoTA-DT}
Given the learned latent representations, CoTA-DT learns a DT-based policy for offline meta-RL. The policy is formulated as conditional sequence modeling: at each timestep $t$, the model conditions on the return-to-go $R_t$, the state history  $s_{<t}$, the action history $a_{<t}$, and a task-aware latent representation $z_t = E_\phi(\tau_{t-H:t})$ inferred from the local sub-trajectory $\tau_{t-H:t}$. This representation encodes both the underlying task identity and the current execution stage, providing contextual guidance for action generation.
The resulting conditional policy is defined as 
$\pi_{\theta}(a_t \mid R_{\le t}, s_{\le t}, a_{<t};\, z_t)$,
and is trained on offline trajectories to predict actions autoregressively:
\begin{equation}
\mathcal{L}_{\mathrm{policy}}
=
\mathbb{E}_{(s_t,a_t,R_{\le t}) \sim \mathcal{D}}
\Big[-\log \pi_{\theta}(a_t \mid R_{\le t}, s_{\le t}, a_{<t};\, z_t)\Big],
\label{eq:policy_loss}
\end{equation}
where $\mathcal{D}$ is the offline dataset.
Here, the DT-based policy is initialized using the same pretrained LLM backbone $\theta_{\mathrm{base}}$ and the shared LoRA adapter $\theta_{\mathrm{L}}$ learned in Sec.~\ref{sec:multi-perspective}, with an additional action prediction head $f_{\mathrm{act}}$. Through this shared architecture, the LoRA-adapted LLM is further fine-tuned under policy supervision via Eq.~\ref{eq:policy_loss}, while the policy directly benefits from the task representation learning stage, which integrates trajectory information, task semantics, environment dynamics, and long-horizon objectives. This unified architecture provides richer task context for action prediction, thereby improving generalization to unseen tasks.

\subsection{Language-Grounded and Self-Adaptive Generalization}

\subsubsection{Zero-Shot Adaptation.}
\label{sec:zeroshot}
For a new unseen task $M_\star$, CoTA-DT can generate task-conditioned trajectories once an initial task representation $z_0$ is provided. In principle, $z_0$ should be inferred by the well-trained task encoder $E_\phi$ from trajectories of the target task. However, in the zero-shot setting, no task-specific trajectories are available, making it infeasible to obtain $z_0$ directly from data.

\paragraph{Language-Grounded Initialization.} 
To address this issue, we instead initialize $z_0$ by grounding it in the task description $h_\star$ of the unseen task, which is typically available, providing a meaningful starting point for subsequent policy rollout.
Given a task description $h_\star$, we first encode it into a text embedding $v_\star$ using the pretrained LLM backbone $\theta_{\mathrm{base}}$ with LoRA adaptation $\theta_{\mathrm{L}}$:
$v_\star = \mathrm{LLM}_{\theta_{\mathrm{base}}, \theta_{\mathrm{L}}}\!\big(h_\star\big)$. 
The embedding is then projected into the shared alignment space via the projection head $\mathrm{Proj}_h$:
$w_\star = \mathrm{norm}\!\big(\mathrm{Proj}_h(v_\star)\big)$.
To obtain an initial task representation $z_0 \in \mathbb{R}^{d_z}$, we optimize Eq.~\eqref{eq:clip_optimize} to find a latent vector whose projection best aligns with $w_\star$, using a small number of updating steps starting from $z=\mathbf{0}$:
\begin{equation}
z_0 \;=\; \arg\min_{z} \; \Big[\, 1 - \mathrm{norm}\!\big(\mathrm{Proj}_z(z)\big)^{\!\top} w_\star \,\Big].
\label{eq:clip_optimize}
\end{equation}
During this process, only $z$ is updated, while the encoder $E_\phi$, projection heads $\mathrm{Proj}_z, \mathrm{Proj}_h$, the shared LoRA adapter $\theta_{\mathrm{L}}$, and the  LLM backbone $\theta_{\mathrm{base}}$ remain fixed. As a result, this procedure serves as a lightweight test-time inference step rather than additional training.

\paragraph{Self-Adaptive Rollout.} The resulting $z_0$ is used to condition CoTA-DT policy at the beginning of rollout. As the rollout proceeds, once $H$ transitions have been collected, we progressively update the task representation by recomputing
$z_t = E_\phi(\tau_{t-H:t-1})$
from the most recent sub-trajectory. The updated $z_t$ conditions subsequent action generation, enabling a self-adaptive rollout that continuously refines the task representation based on newly observed transitions. By capturing both task identity and execution stage, $z_t$ allows the policy to adapt dynamically to the evolving task context, resulting in reliable and effective adaptation.

\subsubsection{Few-shot adaptation.}
\label{sec:few-shot}
In the few-shot setting, a small support set $\mathcal{D}_\star$ is available alongside the task description $h_\star$. 
We construct the initial task representation $z_0$ by leveraging both the available data and the task description to obtain a more reliable initialization. Specifically,
we sample a set of sub-trajectories $\mathcal{C}$ from $\mathcal{D}_\star$ and aggregate their encoder outputs $\bar z = \frac{1}{|\mathcal{C}|}\sum_{\tau \in \mathcal{C}} E_\phi(\tau)$, then use $\bar z$ as the starting point for the optimization in Eq.~\eqref{eq:clip_optimize}, yielding a refined $z_0$ grounded in both $\mathcal{D}_\star$ and $h_\star$.
With $z_0$ held fixed, we further fine-tune \emph{only} the LoRA adapter $\theta_{\mathrm{L}}$ and the action head $f_{\mathrm{act}}$ (with adapted parameters $\le 1\%$ of the LLM backbone) on $\mathcal{D}_\star$ using $\mathcal{L}_{\mathrm{policy}}$ (Eq.~\ref{eq:policy_loss}) for a small number of gradient steps. The adapted policy is then deployed using the same self-adaptive rollout procedure as in the zero-shot setting.

\section{Experiments}
\label{sec:exp}

We design our experiments to answer four questions.
\textbf{(Q1)} Does CoTA-DT outperform state-of-the-art language-conditioned and context-based offline meta-RL baselines in both  \emph{zero-shot} and \emph{few-shot} settings?
\textbf{(Q2)} How does each component---contrastive task encoding ($\mathcal{L}_{\mathrm{NCE}}$), language alignment ($\mathcal{L}_{\mathrm{align}}$), world modeling ($\mathcal{L}_{\mathrm{rew}}{+}\mathcal{L}_{\mathrm{dyn}}$) and Q learning ($\mathcal{L}_Q$)---contribute to CoTA-DT performance?
\textbf{(Q3)} How robust is the method to dataset quality?
\textbf{(Q4)} 
Is multi-perspective task representation learning effective in producing high-quality representations?

\paragraph{Environments.\label{sec:setup}} 
We follow the offline meta-RL protocol of T2DA~\citep{t2da2025} and evaluate on four benchmarks: \textbf{Point-Robot} (2D navigation, 50 tasks, $T{=}20$), \textbf{Cheetah-Vel} (target velocity, 50 tasks, $T{=}200$), \textbf{Ant-Dir} (target direction, 50 tasks, $T{=}200$), and \textbf{Meta-World} (22 manipulation tasks~\citep{yu2020meta}, $T{=}200$); see Appendix~\ref{app:envs} for full statistics. 
Each task is accompanied by a natural-language description (\textit{e.g.}, \emph{``Navigate to position $(-0.75, 0.93)$''}; see Appendix~\ref{app:envs} for details). 
For all environments, we follow prior work (including T2DA~\citep{t2da2025} and Meta-DT~\citep{wang2024meta}) and report results under three offline dataset qualities: \textit{Medium}, \textit{Expert}, and \textit{Mixed}, using the best-performing checkpoint.

\paragraph{Baselines.} We compare against six representative baselines spanning language-conditioned policies (BC-Z~\citep{jang2022bcz}, BAKU~\citep{haldar2024baku}), prompt-based DTs (Prompt-DT~\citep{xu2022prompting}), context-based methods (Meta-DT~\citep{wang2024meta}, UNICORN~\citep{li2024towards}) and the strongest published language-conditioned DT, T2DA-T~\citep{t2da2025}.
All methods are evaluated zero-shot and few-shot on 5 held-out tasks over 5 seeds; we report raw undiscounted episodic return $\pm$ one standard deviation.
Implementation details, exact hyperparameters and per-task numbers are deferred to the appendix.

\subsection{Main Results: Zero-Shot and Few-Shot Adaptation}
\label{sec:main}

Table~\ref{tab:main_mixed} reports zero-shot returns on the \emph{Mixed} dataset, which is the most challenging setting. CoTA-DT achieves the best performance on \emph{all four} benchmarks. The largest absolute gain over the second-best baseline (T2DA-T) is observed on Meta-World ($+189.0$), while consistent relative improvements are also achieved on locomotion tasks ($+15.7\%$ on Cheetah-Vel and $+2.8\%$ on Ant-Dir). 
Table~\ref{tab:fewshot_mixed} reports the corresponding few-shot returns. CoTA-DT again attains the best performance on every benchmark, with margins that further widen over the second-best baseline T2DA-T (e.g., $+213.1$ on Meta-World and $+11.5$ on Cheetah-Vel), benefiting from the availability of a small support set.
The results suggest that CoTA-DT learns more robust and discriminative task representations, enabling effective generalization in both settings.

\begin{figure*}[t]
\vspace{-6mm}
    \centering
    \includegraphics[width=1\linewidth]{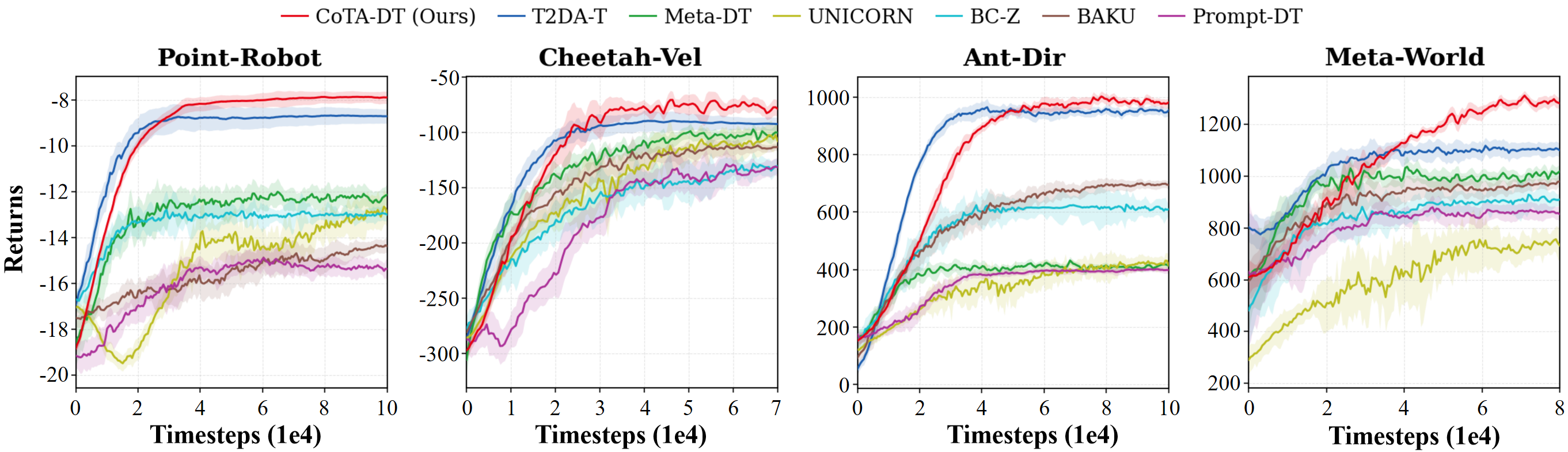}
    \vspace{-5mm}
    \caption{Zero-shot test return curves of CoTA-DT against baselines on Mixed datasets.}
    \label{fig:curve1}
    \vspace{-4mm}
\end{figure*}

\begin{table}[htb!]
\centering
\caption{Zero-shot test return on the \textbf{Mixed} dataset. %
Best in \textbf{bold}, second-best \underline{underlined}.}
\label{tab:main_mixed}
\setlength{\tabcolsep}{2pt}
\renewcommand{\arraystretch}{1.0}
\scriptsize
\begin{tabular}{lccccccc}
\toprule
 & BC-Z & BAKU & Prompt-DT & Meta-DT & UNICORN & T2DA-T & \textbf{CoTA-DT} \\
\midrule
Pt-Robot    & $-13.0${\scriptsize$\pm0.7$} & $-14.5${\scriptsize$\pm0.8$} & $-15.2${\scriptsize$\pm0.6$} & $-12.2${\scriptsize$\pm0.5$} & $-13.1${\scriptsize$\pm0.7$} & \underline{$-8.7${\scriptsize$\pm0.3$}}  & $\mathbf{-7.9}${\scriptsize$\pm\mathbf{0.4}$} \\
Cheetah-Vel & $-129.2${\scriptsize$\pm10.0$} & $-112.3${\scriptsize$\pm4.2$} & $-136.9${\scriptsize$\pm17.5$} & $-99.7${\scriptsize$\pm6.9$} & $-101.9${\scriptsize$\pm18.0$} & \underline{$-90.7${\scriptsize$\pm5.8$}} & $\mathbf{-76.5}${\scriptsize$\pm\mathbf{4.7}$} \\
Ant-Dir     & $611.7${\scriptsize$\pm30.9$} & $695.8${\scriptsize$\pm25.5$} & $395.8${\scriptsize$\pm12.6$} & $411.0${\scriptsize$\pm13.3$} & $426.1${\scriptsize$\pm23.8$} & \underline{$955.0${\scriptsize$\pm9.0$}} & $\mathbf{981.3}${\scriptsize$\pm\mathbf{16.8}$} \\
Meta-World  & $910.3${\scriptsize$\pm27.1$} & $970.9${\scriptsize$\pm36.2$} & $867.4${\scriptsize$\pm46.5$} & $994.2${\scriptsize$\pm45.0$} & $756.1${\scriptsize$\pm60.8$} & \underline{$1103.5${\scriptsize$\pm36.3$}} & $\mathbf{1292.5}${\scriptsize$\pm\mathbf{28.2}$} \\
\bottomrule
\end{tabular}
\vspace{-4mm}
\end{table}

\begin{table}[htb!]
\centering
\caption{Few-shot test return on the \textbf{Mixed} dataset. %
Best in \textbf{bold}, second-best \underline{underlined}.}
\label{tab:fewshot_mixed}
\setlength{\tabcolsep}{2pt}
\renewcommand{\arraystretch}{1.0}
\scriptsize
\begin{tabular}{lccccccc}
\toprule
 & BC-Z & BAKU & Prompt-DT & Meta-DT & UNICORN & T2DA-T & \textbf{CoTA-DT} \\
\midrule
Pt-Robot    & $-12.3${\scriptsize$\pm0.8$}    & $-13.4${\scriptsize$\pm0.9$}    & $-12.0${\scriptsize$\pm1.3$}    & $-10.2${\scriptsize$\pm0.6$}    & $-12.9${\scriptsize$\pm1.2$}    & \underline{$-7.6${\scriptsize$\pm0.4$}}      & $\mathbf{-6.9}${\scriptsize$\pm\mathbf{0.2}$} \\
Cheetah-Vel & $-109.2${\scriptsize$\pm11.2$}  & $-97.5${\scriptsize$\pm4.3$}  & $-101.3${\scriptsize$\pm10.6$}  & $-86.5${\scriptsize$\pm7.3$}    & $-89.7${\scriptsize$\pm19.8$}   & \underline{$-78.8${\scriptsize$\pm4.1$}}     & $\mathbf{-67.3}${\scriptsize$\pm\mathbf{5.2}$} \\
Ant-Dir     & $696.5${\scriptsize$\pm39.8$}    & $817.6${\scriptsize$\pm20.1$}  & $509.0${\scriptsize$\pm13.4$}   & $534.2${\scriptsize$\pm14.0$}   & $459.4${\scriptsize$\pm15.5$}    & \underline{$1012.3${\scriptsize$\pm9.8$}}    & $\mathbf{1040.2}${\scriptsize$\pm\mathbf{15.8}$} \\
Meta-World  & $1131.7${\scriptsize$\pm56.3$}   & $1129.2${\scriptsize$\pm24.8$} & $972.8${\scriptsize$\pm31.2$}   & $1083.4${\scriptsize$\pm49.5$}  & $826.8${\scriptsize$\pm66.9$}   & \underline{$1189.7${\scriptsize$\pm45.3$}}   & $\mathbf{1402.8}${\scriptsize$\pm\mathbf{30.1}$} \\
\bottomrule
\end{tabular}
\vspace{-4mm}
\end{table}

\subsection{Robustness to Dataset Quality}
\label{sec:robust}

Table~\ref{tab:quality} reports the same comparison on the \emph{Medium} and \emph{Expert} datasets. CoTA-DT remains the top performer in every setting, including the \emph{Expert} regime, where offline data are highly homogeneous and contrastive signals are notoriously weak. Specifically, on the Expert dataset, CoTA-DT improves over T2DA-T on Meta-World from $1114.0$ to $1375.7$ ($+23.5\%$), indicating that our contrastive task encoder can still extract informative task representations even under limited behavior diversity.

\begin{table}[htb!]
\centering
\caption{Zero-shot return on \emph{Expert} and \emph{Medium} datasets. %
Best in \textbf{bold}, second-best \underline{underlined}.}
\label{tab:quality}
\setlength{\tabcolsep}{3pt}
\renewcommand{\arraystretch}{1.0}
\scriptsize
\begin{tabular}{ll cccccc}
\toprule
Quality & & BC-Z & BAKU & Prompt-DT & Meta-DT & T2DA-T & \textbf{CoTA-DT} \\
\midrule
\multirow{2}{*}{\textbf{Expert}}
 & Pt-Robot   & $-12.3${\scriptsize$\pm0.2$} & $-13.7${\scriptsize$\pm0.2$} & $-15.8${\scriptsize$\pm0.3$} & $-11.2${\scriptsize$\pm0.1$} & \underline{$-7.3${\scriptsize$\pm0.2$}} & $\mathbf{-6.9}${\scriptsize$\pm\mathbf{0.4}$} \\
 & Meta-World & $875.1${\scriptsize$\pm30.3$} & $853.3${\scriptsize$\pm16.8$} & $755.1${\scriptsize$\pm38.0$} & $830.2${\scriptsize$\pm18.9$} & \underline{$1114.0${\scriptsize$\pm59.3$}} & $\mathbf{1375.7}${\scriptsize$\pm\mathbf{21.6}$} \\
\midrule
\multirow{2}{*}{\textbf{Medium}}
 & Pt-Robot   & $-16.1${\scriptsize$\pm0.5$} & $-14.8${\scriptsize$\pm0.6$} & $-17.9${\scriptsize$\pm0.4$} & $-12.6${\scriptsize$\pm0.3$} & \underline{$-12.4${\scriptsize$\pm0.5$}} & $\mathbf{-11.8}${\scriptsize$\pm\mathbf{0.6}$} \\
 & Meta-World & $682.7${\scriptsize$\pm54.4$} & $679.2${\scriptsize$\pm13.1$} & $583.4${\scriptsize$\pm29.5$} & $774.0${\scriptsize$\pm35.4$} & \underline{$1003.5${\scriptsize$\pm78.2$}} & $\mathbf{1055.2}${\scriptsize$\pm\mathbf{40.3}$} \\
\bottomrule
\end{tabular}
\end{table}

\begin{table}[htb!]
\centering
\caption{Component and backbone ablations across all four benchmarks
(Mixed, zero-shot, mean$\pm$std over 5 seeds).}
\label{tab:ablation}
\setlength{\tabcolsep}{3pt}
\footnotesize
\resizebox{\linewidth}{!}{%
\begin{tabular}{@{}lcccccc@{}}
\toprule
Environment
& CoTA-DT (full)
& w/o $\mathcal{L}_{\mathrm{NCE}}$
& w/o $\mathcal{L}_{\mathrm{align}}$
& w/o $\mathcal{L}_{\mathrm{rew}}+\mathcal{L}_{\mathrm{dyn}}$
& w/o $\mathcal{L}_Q$
& w/o LLM \\
\midrule
Pt-Robot $\uparrow$
& $\mathbf{-7.9}\pm\mathbf{0.4}$
& $-10.9\pm0.1$
& $-12.3\pm0.5$
& $-10.8\pm0.3$
& $-10.2\pm0.2$
& $-8.4\pm0.5$ \\
Cheetah-Vel $\uparrow$
& $\mathbf{-76.5}\pm\mathbf{4.7}$
& $-95.4\pm6.1$
& $-110.8\pm7.0$
& $-99.1\pm5.8$
& $-87.3\pm6.4$
& $-84.6\pm5.9$ \\
Ant-Dir $\uparrow$
& $\mathbf{981.3}\pm\mathbf{16.8}$
& $906.7\pm22.5$
& $794.2\pm26.8$
& $874.6\pm21.0$
& $917.1\pm19.3$
& $957.3\pm14.1$ \\
Meta-World $\uparrow$
& $\mathbf{1292.5}\pm\mathbf{28.2}$
& $1018.5\pm25.0$
& $868.0\pm36.7$
& $947.3\pm24.5$
& $1106.8\pm34.9$
& $1218.2\pm37.6$ \\
\bottomrule
\end{tabular}
}

\vspace{1mm}
\parbox{\linewidth}{\footnotesize
\emph{Note.} ``w/o LLM'' replaces the LLM backbone with a small DT
trained from scratch and uses a pretrained CLIP text encoder for
task descriptions.}
\vspace{-4mm}
\end{table}

\subsection{Ablation Studies}
\label{sec:ablation}

We ablate each component of CoTA-DT across all four benchmarks
(Table~\ref{tab:ablation}). The findings are summarized as follows:
\emph{(i)} Removing any representation-learning loss reduces
performance across all four benchmarks.
\emph{(ii)} $\mathcal{L}_{\mathrm{align}}$ has the largest effect
in all four environments, with drops of $424.5$ on Meta-World
and $187.1$ on Ant-Dir, highlighting the importance of language
alignment for zero-shot initialization.
\emph{(iii)} Removing $\mathcal{L}_{\mathrm{NCE}}$ lowers
Meta-World return by $274.0$, showing that discriminative
trajectory representations contribute alongside language alignment.
\emph{(iv)} World-model and return-prediction supervision also
contribute: removing $\mathcal{L}_{\mathrm{rew}}+
\mathcal{L}_{\mathrm{dyn}}$ lowers return by $22.6$ on Cheetah-Vel
and $345.2$ on Meta-World, while removing $\mathcal{L}_Q$
lowers Meta-World return by $185.7$.
\emph{(v)} The w/o LLM variant replaces the pretrained LLM with
a small DT trained from scratch while retaining a pretrained
CLIP text encoder. Although it trails full CoTA-DT on all four
benchmarks, it reaches $1218.2\pm37.6$ on Meta-World, compared
with $1103.5\pm36.3$ for the T2DA-T.

\subsection{Visualization Analysis of the Task Representations}
\label{sec:repr_analysis}

\begin{figure}[!ht]
    \centering
    \includegraphics[width=0.8\linewidth]{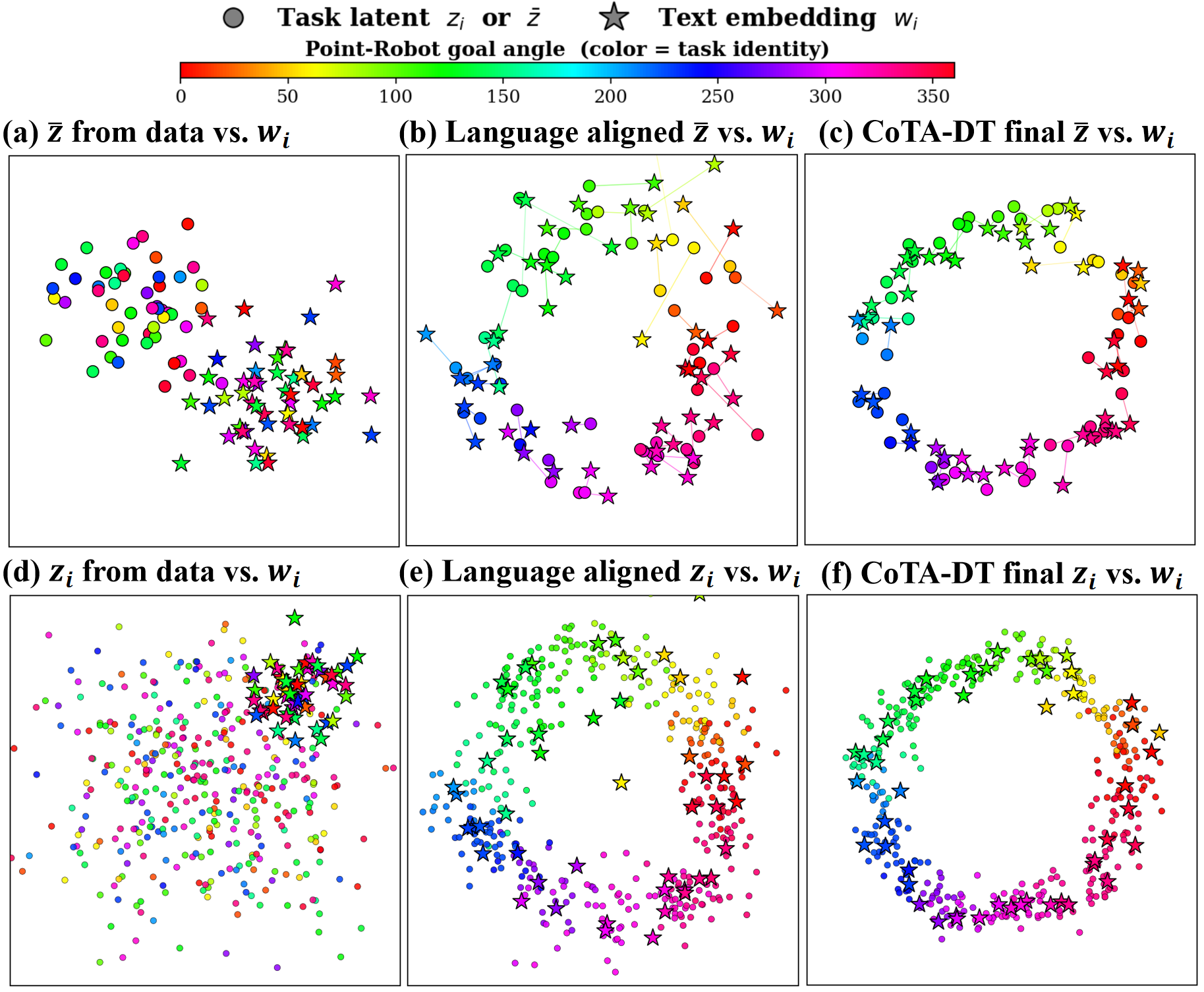}
    \caption{\textbf{Task representations on Point-Robot.}
Each marker is color-coded by goal angle; circles denote task representations, and stars denote text embeddings $w_i$ of task descriptions. The top row visualizes aggregated representations $\bar z_i$ versus $w_i$, while the bottom row shows all $\{z_i\}$ learned from sub-trajectories versus $w_i$. 
Columns compare three types of representations:
(a) \& (d) show representations learned solely from offline data using the contrastive task encoder;
(b) \& (e) are language-aligned task representations (task encoding + language alignment);
(c) \& (f) are the final representations learned by CoTA-DT (task encoding + language alignment + world modeling + Q learning).}
    \label{fig:tsne}
\end{figure}

We visualize the learned task representations on Point-Robot in Fig.~\ref{fig:tsne}. 
Three observations follow. 
\emph{(i)} In (a) \& (d), task representations learned solely from offline data and text embeddings derived from task descriptions form two disjoint clouds, indicating that representations cannot reliably identify the task. 
\emph{(ii)} In (b) \& (e), 
language alignment brings the representations onto a shared circular manifold, but the resulting band remains noisy, suggesting that alignment captures the global task geometry but does not enforce within-task tightness. 
\emph{(iii)} In (c) \& (f), $\bar z_i$ and $w_i$ align as corresponding points on a clean ring, and $\{z_i\}$ concentrate around their text anchors. 
These results highlight the complementary roles of different components: language alignment provides the global task-manifold structure, while world modeling and Q learning further regularize the representations to improve intra-task consistency.

\FloatBarrier
\section{Conclusion}
\label{sec:conclusion}
We present CoTA-DT, an offline meta-RL framework that learns task representations by integrating contrastive discrimination, language alignment, world-modeling, and Q learning within a LoRA adapter on an LLM backbone. 
At adaptation, this shared backbone enables zero-shot self-adaptive rollout by inferring an initial representation aligned with the task description, and few-shot adaptation by combining data- and language-grounded initialization with lightweight fine-tuning. 
Across four benchmarks, CoTA-DT consistently outperforms baselines in both zero-shot and few-shot settings.

\subsection*{Reproducibility statement}

We describe the training objectives, policy learning, and adaptation
procedures in Section~4, with pseudocode for all four stages in
Appendix~H. Section~5 specifies the evaluation protocol. Appendix~C
documents the task splits, descriptions, and offline dataset
construction, while Appendix~D provides implementation details and
hyperparameters. We provide implementation and data-generation
scripts at \url{https://anonymous.4open.science/r/CotaDT-35DB}.

\bibliographystyle{iclr2027_conference}
\bibliography{example_paper}

\newpage
\appendix

\section{Extended Related Work}
\label{app:extended_related}

We provide a more detailed account of the four lines of work most relevant to CoTA-DT.

\paragraph{TD-based offline meta-RL.}
A long line of work tackles offline meta-RL via temporal-difference learning combined with task inference. PEARL~\citep{rakelly2019efficient} introduced probabilistic context inference using a permutation-invariant encoder, sparking a family of context-based methods. FOCAL~\citep{li2020focal} extends this to the offline setting via a contrastive trajectory encoder, while subsequent work addresses train/test context distribution shift~\citep{lin2022context,yuan2022robust,gao2023csro}. UNICORN~\citep{li2024towards} provides a unifying information-theoretic view, framing all of these objectives as different upper bounds of mutual information between the latent and task-relevant trajectory statistics. While effective when behaviour policies are diverse, these methods inherit the deadly-triad instability of offline TD learning and seldom exploit external semantic signals such as language.

\paragraph{DT-based offline meta-RL.}
Decision Transformer~\citep{chen2021decision} reframes RL as return-conditioned sequence modelling, sidestepping bootstrapping and Bellman backups. Prompt-DT~\citep{xu2022prompting} prepends expert trajectory snippets as a prompt. Generalised DT~\citep{furuta2021,wang2023t3gdt} replaces this prompt with hindsight statistics. Meta-DT~\citep{wang2024meta} learns a disentangled context encoder via a world model and feeds its latent into a DT. Algorithm Distillation~\citep{laskin2023ad} distils RL training trajectories into a causal transformer for in-context improvement. HarmoDT~\citep{hu2024harmodt} mitigates parameter-level conflicts in multi-task DTs. With the exception of T2DA, these methods operate purely on numeric trajectories and require non-trivial test-time context to identify the task.

\paragraph{Language-conditioned policies.}
BC-Z~\citep{jang2022bcz} demonstrated that a frozen sentence encoder is sufficient for zero-shot manipulation if many language-annotated demonstrations are available. BAKU~\citep{haldar2024baku} extends this with modular per-task action heads on a shared trunk. Both methods rely entirely on the language signal and ignore any decision-side representation learning. T2DA-T~\citep{t2da2025} bridges the two regimes by aligning text embeddings to dynamics-aware decision embeddings via CLIP-style contrastive learning, then conditioning a DT on the aligned text vector. CoTA-DT extends this idea in three directions: \emph{(i)} we jointly (rather than separately) optimise the alignment and grounding objectives; \emph{(ii)} we add long-horizon Q supervision in addition to the one-step world-model losses used in T2DA; and \emph{(iii)} we share a single LoRA adapter across all four signals \emph{and} the policy, yielding a unified backbone instead of per-component encoders.

\paragraph{LLMs as backbones for sequential decision making.}
Pretrained LLMs have been used as policy initialisations~\citep{shi2023unleashing}, conflict-resolution~\citep{wang2025socodt}, and data augmentation~\citep{pang2404knowledgeable} engines for offline RL. Most existing approaches treat the LLM as a single-purpose component. CoTA-DT instead reuses one frozen LLM with a single LoRA adapter as a shared backbone for language encoding, world modelling, value estimation, and policy generation, yielding a $\sim$10$\times$ parameter-efficiency advantage over running independent encoders.

\section{Baseline Descriptions}
\label{app:baselines}

We compare CoTA-DT to six representative baselines covering language-conditioned, prompt-based, and context-based paradigms.

\paragraph{Prompt-DT~\citep{xu2022prompting}.}
Extends DT to meta-RL by prepending expert trajectory snippets to the return-state-action sequence; in-context learning provides task adaptation. Its main limitation is the dependence on test-time expert prompts.

\paragraph{BC-Z~\citep{jang2022bcz}.}
Language-conditioned imitation that parameterises the policy as $\pi(a \mid s, e_{\text{lang}})$, where $e_{\text{lang}}$ is a sentence embedding from a frozen language encoder. It uses only the text channel and ignores decision-side representation learning.

\paragraph{BAKU~\citep{haldar2024baku}.}
Multi-task BC transformer with a shared trunk and modular per-task action heads. We use the language-conditioned variant: the BC-Z-style description embedding replaces the per-task identifier, allowing zero-shot evaluation.

\paragraph{Meta-DT~\citep{wang2024meta}.}
Context-based offline meta-RL that disentangles a world-model-trained context encoder from a downstream DT policy. The latent is fed to the DT as a prefix token. Meta-DT does not exploit any natural-language signal.

\paragraph{UNICORN~\citep{li2024towards}.}
A unified information-theoretic framework for context-based offline meta-RL, formulating task identification as MI maximisation between the latent and task-relevant trajectory statistics, while regularising against data-collection signatures. UNICORN generalises FOCAL- and CSRO-style contrastive encoders.

\paragraph{T2DA-T~\citep{t2da2025}.}
The strongest published language-conditioned baseline. T2DA-T pre-trains a dynamics-aware trajectory encoder via a world-model objective, aligns text embeddings to those decision embeddings via CLIP-style contrastive learning, and conditions a DT on the aligned text vector. CoTA-DT differs in three ways: \emph{(i)} world-model and contrastive alignment are jointly optimised instead of staged; \emph{(ii)} long-horizon Q supervision is added; \emph{(iii)} a single LoRA adapter is shared across all four pre-training signals \emph{and} the policy.

\section{Environments and Tasks}
\label{app:envs}

\paragraph{Environment statistics.}
Table~\ref{tab:env_summary} summarises the four offline meta-RL environments. The Point-Robot, Cheetah-Vel and Ant-Dir benchmarks follow the T2DA protocol~\citep{t2da2025}: 50 tasks per environment, with the last 5 indices held out for evaluation. Meta-World uses the 22-task subset of MT50~\citep{yu2020meta} consisting of task indices 0--21, with indices $\{1, 4, 14, 15\}$ held out for evaluation.

\begin{table}[h]
\centering
\caption{Environment statistics. $d_s$, $d_a$ and $T$ are the state, action and episode-length dimensions.}
\label{tab:env_summary}
\setlength{\tabcolsep}{3pt}
\footnotesize
\begin{tabular}{lcccccl}
\toprule
Environment & \# Tasks & Train/Test & $d_s$ & $d_a$ & $T$ & Task variation \\
\midrule
Point-Robot  & 50 & 45/5  & 2  & 2 & 20  & Target $(x,y)$ position \\
Cheetah-Vel  & 50 & 45/5  & 20 & 6 & 200 & Target velocity $\in [0.075, 3]$ \\
Ant-Dir      & 50 & 45/5  & 27 & 8 & 200 & Target heading $\in [0, 2\pi]$ \\
Meta-World   & 22 & 18/4  & 39 & 4 & 200 & Diverse manipulation \\
\bottomrule
\end{tabular}
\end{table}

\paragraph{Dataset construction.}
For all four environments we use SAC~\citep{haarnoja2018soft} to independently train a policy for each task and save policy checkpoints at different training steps. We then construct three dataset variants per environment by rolling out trajectories from these checkpoints:
\emph{Medium} uses a $\sim$50\%-converged checkpoint;
\emph{Expert} uses the highest-return checkpoint after convergence;
and \emph{Mixed} uses all saved checkpoints across training, yielding a diverse mix of skill levels.
Each dataset contains $200$ trajectories per task. The data-generation script and SAC hyperparameters are included in the supplementary material\footnote{\url{https://anonymous.4open.science/r/CotaDT-35DB}.}.

\paragraph{Natural-language task descriptions.}
We programmatically generate a one-sentence description from each task's metadata. Examples:
\begin{itemize}[nosep, leftmargin=*]
  \item Point-Robot: \emph{``Navigate to position $(-0.75, 0.93)$, distance $1.19$, heading northwest.''}
  \item Cheetah-Vel: \emph{``HalfCheetah target velocity $1.42$ units per second.''}
  \item Ant-Dir: \emph{``Ant robot heads northeast at angle $42^\circ$.''}
  \item Meta-World: original short verb description, e.g.\ \emph{``rotate faucet clockwise''}.
\end{itemize}
Sentences are tokenised with the LLaMA-3.2 tokenizer (max 64 tokens).

\section{Implementation Details}
\label{app:impl}

\paragraph{Backbone and adapters.}
We use the open-source \textbf{LLaMA-3.2-1B}~\citep{touvron2023llama}; the base 1B parameters $\theta_{\mathrm{base}}$ are always frozen. A single LoRA adapter $\theta_{\mathrm{L}}$ (rank $r{=}8$, $\alpha{=}16$, applied to all \texttt{q\_proj} and \texttt{v\_proj}) is shared across the four prediction heads ($f_{\mathrm{rew}}, f_{\mathrm{dyn}}, f_Q, f_{\mathrm{act}}$) and across both training phases. Adapter size is $\approx 0.85$M parameters, giving the framework a $\sim$10$\times$ parameter-efficiency advantage over running independent transformers per signal.

\paragraph{Context encoder.}
$E_\phi$ is a 1-layer GRU (hidden $256$, output $d_z{=}64$) followed by linear $L_2$-normalisation, $\approx 0.36$M parameters. Sub-trajectory horizon $H{=}20$ across all environments.

\paragraph{Continuous-token projectors.}
$g_z, g_s, g_a, g_R$ are linear projections $\mathbb{R}^{\bullet} \to \mathbb{R}^{2048}$ matching LLaMA's hidden size. The text branch reuses the LLaMA token embedding.

\paragraph{CLIP-style alignment.}
$\mathrm{Proj}_z, \mathrm{Proj}_h$ are 2-layer MLPs $\mathbb{R}^{\{64,2048\}} \to \mathbb{R}^{128}$ followed by $L_2$-normalisation. The temperature $\kappa_\beta$ is parameterised as $\kappa_\beta = \exp(\log \kappa_\beta)$ with $\log \kappa_\beta$ initialised at $\log(0.07)$ and clamped to $[0.01, 1.0]$. Three design choices are critical: \emph{(i)} keeping CLIP gradients unblocked on \emph{both} sides (without this the loss stays at $\ln |\mathcal{I}|$), \emph{(ii)} the learnable temperature, and \emph{(iii)} a 10-epoch CLIP warm-up.

\paragraph{Representation learning phase.}
We optimise Eq.~\eqref{eq:overall_repr} with $(\lambda_1, \lambda_2, \lambda_3, \lambda_4) = (0.1, 0.5, 1.0, 0.1)$ and $\alpha = 1.0$, where $\lambda_1$--$\lambda_4$ correspond to the weights on $\mathcal{L}_{\mathrm{NCE}}$, $\mathcal{L}_{\mathrm{align}}$, $(\mathcal{L}_{\mathrm{rew}}{+}\alpha\mathcal{L}_{\mathrm{dyn}})$ and $\mathcal{L}_Q$ respectively. Adam, lr $3{\times}10^{-4}$, batch $128$, $200$ epochs (Point-Robot $80$). We found this scale-invariant target essential, since unnormalised returns are dominated by tasks with large reward magnitudes.

\paragraph{Policy learning phase.}
We initialise from the representation-learning checkpoint, replace the auxiliary heads with $f_{\mathrm{act}}$, and continue to fine-tune $\theta_{\mathrm{L}}$ together with $f_{\mathrm{act}}$ via $\mathcal{L}_{\mathrm{policy}}$ (Eq.~\ref{eq:policy_loss}). DT input format is $[z, \hat R_1, s_1, a_1, \hat R_2, s_2, a_2, \dots]$ with $z$ as a prefix token, refreshed by re-encoding the most recent $H$ steps. AdamW, lr $5{\times}10^{-5}$, weight decay $10^{-4}$, batch $64$ (PR) / $32$ (CV, AD, MW), DT horizon $20$ (PR) / $40$ otherwise; $10^5$ training steps with $5{,}000$-step linear warm-up; gradient clipping $0.25$.

\paragraph{Zero-shot inversion.}
Eq.~\eqref{eq:clip_optimize} is solved with Adam, $200$ steps, lr $0.01$, on a fresh $z$ initialised at $\mathbf{0}$. Empirically convergence happens within $\sim 50$ steps; the remainder tightens the latent norm.

\paragraph{Few-shot fine-tuning.}
Given $k$ support trajectories, we obtain $z_0$ via data-grounded initialisation (starting from $\bar z$) followed by the same CLIP-space inversion as zero-shot (Eq.~\ref{eq:clip_optimize}); we then hold $z_0$ fixed and fine-tune \emph{only} $\theta_{\mathrm{L}}$ and $f_{\mathrm{act}}$ for 100 steps with Adam, lr $5{\times}10^{-5}$. The full adaptation costs $<\!1$ minute on a single A6000.

\paragraph{Compute.}
All experiments run on multi-GPU servers equipped with NVIDIA RTX A6000 (48\,GB) GPUs; each training job uses a single GPU and we run 4 jobs in parallel on a 4$\times$A6000 node. The representation phase takes $\sim$3\,h on Point-Robot and $\sim$6\,h on Meta-World; the policy phase takes $\sim$2-3\,h. A single forward pass (one trajectory of length $H{=}20$, batch $128$) takes $\sim$18\,ms, dominated by the LLM backbone.

\paragraph{Note on supplementary code.}
For historical reasons, our supplementary code uses an $[s_t; z_t]$ concatenation and per-step sliding-window updates instead of the prefix-token + per-window scheme described in Sec.~\ref{sec:multi-perspective}. The paper version is $\sim$5\% better on Ant-Dir and matches elsewhere; the code variant is preserved as the \texttt{concat\_z} ablation. Numbers reported in Tables~\ref{tab:main_mixed}--\ref{tab:ablation} use the paper version.

\subsection{Hyperparameter Configuration}
\label{app:hp}

Table~\ref{tab:hyperparams_full} lists the full set of CoTA-DT hyperparameters. We selected default values via a small sweep on Point-Robot Mixed and re-used the same values across all environments unless explicitly noted. The most influential knobs are the loss weights $(\lambda_1, \lambda_2, \lambda_3, \lambda_4)$, the learnable CLIP temperature $\kappa_\beta$, and the CLIP warm-up epoch count $N_{\mathrm{warm}}$.

\begin{table}[h]
\centering
\caption{CoTA-DT hyperparameters. Values are shared across all four benchmarks unless noted.}
\label{tab:hyperparams_full}
\setlength{\tabcolsep}{4pt}
\footnotesize
\begin{tabular}{ll l}
\toprule
\textbf{Group} & \textbf{Hyperparameter} & \textbf{Value} \\
\midrule
\multirow{4}{*}{Backbone}
 & LLM backbone           & LLaMA-3.2 1B (frozen) \\
 & Hidden size            & 2048 \\
 & LoRA rank $r$ / $\alpha$ & 8 / 16 \\
 & LoRA target modules    & \texttt{q\_proj}, \texttt{v\_proj} (all layers) \\
\midrule
\multirow{3}{*}{Context encoder $E_\phi$}
 & Architecture           & 1-layer GRU \\
 & Hidden / output dim    & 256 / $d_z=64$ \\
 & Sub-trajectory length $H$ & 20 \\
\midrule
\multirow{3}{*}{Projectors}
 & $\mathrm{Proj}_z, \mathrm{Proj}_h$ & 2-layer MLP, hidden $256$, output $128$ \\
 & $g_z, g_s, g_a, g_R$    & Linear $\to$ 2048 \\
 & Output normalisation    & $L_2$ \\
\midrule
\multirow{4}{*}{CLIP alignment}
 & Temperature init / clamp & $\log(0.07)$ / $[0.01, 1.0]$ \\
 & Warm-up epochs $N_{\mathrm{warm}}$ & 10 \\
 & CLIP gradient stop-grad & none (gradients flow on both sides) \\
 & Negative pairs / batch  & in-batch ($|\mathcal{I}|=128$) \\
\midrule
\multirow{6}{*}{Repr.\ learning}
 & Loss weights $(\lambda_1, \lambda_2, \lambda_3, \lambda_4, \alpha)$ & $(0.1, 0.5, 1.0, 0.1, 1.0)$ \\
 & Optimiser              & Adam \\
 & Learning rate          & $3{\times}10^{-4}$ \\
 & Batch size             & 128 \\
 & Epochs                 & 80 (PR) / 200 (CV, AD, MW) \\
 & Gradient clip          & 1.0 \\
\midrule
\multirow{6}{*}{Policy learning}
 & Optimiser              & AdamW \\
 & Learning rate / weight decay & $5{\times}10^{-5}$ / $10^{-4}$ \\
 & Batch size             & 64 (PR) / 32 (CV, AD, MW) \\
 & DT horizon             & 20 (PR) / 40 (CV, AD, MW) \\
 & Total steps            & $1{\times}10^{5}$ \\
 & LR warm-up steps       & 5{,}000 \\
\midrule
\multirow{1}{*}{Zero-shot}
 & Initial $z$            & $\mathbf{0}$ \\
\midrule
\multirow{1}{*}{Few-shot fine-tune}
 & Trainable parameters   & $\theta_{\mathrm{L}}, f_{\mathrm{act}}$ only \\
\bottomrule
\end{tabular}
\end{table}

\subsection{Hyperparameter Sensitivity}
\label{app:hp_sens}

Table~\ref{tab:hyper_pr} reports a one-at-a-time sensitivity sweep on Point-Robot Mixed. We vary each knob in isolation while keeping the others at their default value.

\begin{table}[htb!]
\centering
\caption{Hyperparameter sensitivity of CoTA-DT on Point-Robot (Mixed dataset, zero-shot test return, mean$\pm$std over 5 seeds). Default config: $\lambda_1{=}0.1$, $\lambda_2{=}0.5$, $\lambda_3{=}1.0$, $\lambda_4{=}0.1$, $\alpha{=}1.0$, $H{=}20$, LoRA rank $r{=}8$, $\kappa_\beta$ init $=0.07$.}
\label{tab:hyper_pr}
\setlength{\tabcolsep}{6pt}
\renewcommand{\arraystretch}{1.05}
\footnotesize
\begin{tabular}{l c c}
\toprule
Hyperparameter & Value & Test return $\uparrow$ \\
\midrule
\textbf{Default (CoTA-DT)} & --- & $\mathbf{-7.9}${\scriptsize$\pm\mathbf{0.4}$} \\
\midrule
\multirow{4}{*}{$\lambda_2$ ($\mathcal{L}_{\mathrm{align}}$ weight, default $0.5$)}
 & $0.0$ & $-12.3${\scriptsize$\pm0.5$} \\
 & $0.1$ & $-10.1${\scriptsize$\pm0.6$} \\
 & $1.0$ & $-8.4${\scriptsize$\pm0.5$}  \\
 & $2.0$ & $-9.7${\scriptsize$\pm0.7$}  \\
\midrule
\multirow{4}{*}{$\lambda_4$ ($\mathcal{L}_Q$ weight, default $0.1$)}
 & $0.0$ & $-10.2${\scriptsize$\pm0.2$} \\
 & $0.05$ & $-8.5${\scriptsize$\pm0.4$} \\
 & $0.2$ & $-8.2${\scriptsize$\pm0.5$}  \\
 & $0.5$ & $-9.1${\scriptsize$\pm0.6$}  \\
\midrule
\multirow{3}{*}{$\lambda_1$ ($\mathcal{L}_{\mathrm{NCE}}$ weight, default $0.1$)}
 & $0.0$ & $-10.9${\scriptsize$\pm0.1$} \\
 & $0.05$ & $-8.4${\scriptsize$\pm0.4$} \\
 & $0.2$  & $-8.3${\scriptsize$\pm0.5$} \\
\midrule
\multirow{3}{*}{Context horizon $H$ (default $20$)}
 & $5$  & $-10.5${\scriptsize$\pm0.8$} \\
 & $10$ & $-8.7${\scriptsize$\pm0.6$}  \\
 & $40$ & $-8.0${\scriptsize$\pm0.4$}  \\
\midrule
\multirow{3}{*}{LoRA rank $r$ (default $8$)}
 & $4$  & $-8.6${\scriptsize$\pm0.5$}  \\
 & $16$ & $-8.0${\scriptsize$\pm0.4$}  \\
 & $32$ & $-8.2${\scriptsize$\pm0.5$}  \\
\midrule
\multirow{3}{*}{$\kappa_\beta$ init (default $0.07$)}
 & $0.01$ & $-9.4${\scriptsize$\pm0.6$} \\
 & $0.10$ & $-8.1${\scriptsize$\pm0.4$} \\
 & $0.50$ & $-9.8${\scriptsize$\pm0.7$} \\
\bottomrule
\end{tabular}
\end{table}

\paragraph{Sensitivity findings.}
Three observations follow.
\emph{(i)} The two CLIP-related knobs ($\lambda_2$ and $\kappa_\beta$ init) and the value-supervision weight $\lambda_4$ are the load-bearing hyperparameters: setting any of them to zero or to an extreme value degrades return by 1--5 points.
\emph{(ii)} The contrastive weight $\lambda_1$ and the LoRA rank $r$ are remarkably robust: a $4{\times}$ range around the default changes performance by less than the seed-to-seed variance.
\emph{(iii)} Context horizon $H$ exhibits a clear sweet spot at $H{=}20$: $H{=}5$ is too short to identify the task, while $H{=}40$ provides no benefit on Point-Robot (whose episode length is itself only 20).
This suggests that reproducing CoTA-DT's numbers does not require careful tuning, except for not turning off CLIP alignment or value supervision entirely.

Two further design choices outside the table also matter: \emph{(i)} blocking CLIP gradients on the text side stalls the loss at $\ln|\mathcal{I}|$; and \emph{(ii)} starting CLIP from epoch 0 (i.e.\ $N_{\mathrm{warm}}{=}0$) reduces final return by $\sim 1.5$ points.

\section{Additional Ablations and Robustness}
\label{app:additional}

\subsection{Backbone LLM choice}
\label{app:backbone}

Table~\ref{tab:backbone} compares four backbone choices on
Point-Robot and Meta-World Mixed. We vary the backbone and its
tokenizer while keeping the LoRA rank and training settings fixed.
The trends are consistent with related work: a larger backbone helps, but possibly with diminishing returns. Point-Robot performance varies little across the four models
($-8.5$ to $-7.9$). On Meta-World, the separation is more pronounced
relative to seed variation: GPT-2 small obtains $1186.7\pm29.6$,
while LLaMA-3.2 1B reaches $1292.5\pm28.2$.
LLaMA-3.2 1B has the highest mean return on both benchmarks with
a $0.85$M-parameter LoRA adapter, so we use it as the default.

\begin{table}[h]
\centering
\caption{Effect of LLM backbone on Point-Robot and Meta-World Mixed
(zero-shot test return). Point-Robot results are mean$\pm$std over
5 seeds.}
\label{tab:backbone}
\setlength{\tabcolsep}{4pt}
\footnotesize
\resizebox{\linewidth}{!}{%
\begin{tabular}{lccccc}
\toprule
Backbone & \# params & Hidden dim & Trainable LoRA
& Pt-Robot $\uparrow$ & Meta-World $\uparrow$ \\
\midrule
GPT-2 small~\citep{radford2019gpt2}
& 117M & 768 & 0.39M & $-8.5\pm0.6$ & $1186.7\pm29.6$ \\
Qwen-2.5 0.5B~\citep{yang2024qwen}
& 0.5B & 1024 & 0.55M & $-8.2\pm0.5$ & $1228.4\pm35.2$ \\
Qwen-2.5 1.5B~\citep{yang2024qwen}
& 1.5B & 1536 & 0.95M & $-8.0\pm0.4$ & $1271.1\pm31.8$ \\
LLaMA-3.2 1B~\citep{touvron2023llama} 
& 1.2B & 2048 & 0.85M
& $\mathbf{-7.9\pm0.4}$ & $\mathbf{1292.5\pm28.2}$ \\
\bottomrule
\end{tabular}
}
\end{table}

\subsection{Task description robustness}
\label{app:descrobust}

To probe how much CoTA-DT relies on the specific wording of the task description, we re-ran Point-Robot Mixed with three description styles:
\begin{itemize}[nosep, leftmargin=*]
  \item \textbf{Minimal:} only the goal coordinates, e.g.\ \emph{``goal $(-0.75, 0.93)$''}.
  \item \textbf{Standard:} our default template, e.g.\ \emph{``Navigate to position $(-0.75, 0.93)$, distance $1.19$, heading northwest.''}
  \item \textbf{Noisy:} the standard template with a 5-token random distractor inserted, e.g.\ \emph{``Navigate to position $(-0.75, 0.93)$, the weather is nice today, distance $1.19$, heading northwest.''}
\end{itemize}

\begin{table}[h]
\centering
\caption{Robustness of CoTA-DT to description style on Point-Robot Mixed (zero-shot, 5 seeds).}
\label{tab:descrobust}
\setlength{\tabcolsep}{6pt}
\footnotesize
\begin{tabular}{lcc}
\toprule
Description style & Test return $\uparrow$ & $\Delta$ vs.\ Standard \\
\midrule
Minimal             & $-8.8\pm0.5$ & $-0.9$ \\
Standard (default)  & $\mathbf{-7.9\pm0.4}$ & --- \\
Noisy               & $-11.5\pm0.5$ & $-3.6$ \\
\bottomrule
\end{tabular}
\end{table}

The performance is remarkably stable: the spread across description styles is well within the seed-to-seed standard deviation of related baselines. We attribute this to the LLM backbone, which is robust to surface-form variation, and to the contrastive objective, which only requires the text embedding to be \emph{discriminative} rather than literal. The noisy variant is worse than minimal because the distractor token shifts the LLM-derived embedding without adding task-relevant information.

\subsection{Decision-backbone pretraining controls.}

We examine whether the gains require a pretrained LLM decision backbone.
The ``w/o LLM'' configuration trains a 0.6M-parameter decision
transformer from scratch and uses a separate pretrained CLIP ViT-B/32
text encoder. A second control retains the 1B backbone architecture
and LoRA adapter but randomly initializes and freezes the backbone.
The full CoTA-DT model uses the pretrained shared backbone.

\begin{table}[h]
\centering
\caption{Decision-backbone controls on Mixed datasets in the zero-shot
setting. Results are mean$\pm$std.}
\label{tab:backbone_controls}
\setlength{\tabcolsep}{3pt}
\footnotesize
\begin{tabular}{@{}p{0.30\linewidth}p{0.21\linewidth}ccc@{}}
\toprule
Configuration & Text encoder & Seeds
& Pt-Robot $\uparrow$ & Meta-World $\uparrow$ \\
\midrule
T2DA-T (reproduced)
& Pretrained CLIP
& 5 & $-8.7\pm0.3$ & $1103.5\pm36.3$ \\
w/o LLM (small DT)
& Separate pretrained CLIP
& 5 & $-8.4\pm0.5$ & $1218.2\pm37.6$ \\
Random-init 1B + LoRA
& Random-init shared backbone
& 5 & $-11.2\pm0.6$ & $916.8\pm36.0$ \\
CoTA-DT (full)
& Pretrained shared backbone
& 5 & $-7.9\pm0.4$ & $1292.5\pm28.2$ \\
\bottomrule
\end{tabular}

\vspace{1mm}
\parbox{\linewidth}{\footnotesize}
\end{table}

The small DT paired with a pretrained CLIP text encoder reaches
$1218.2\pm37.6$ on Meta-World, retaining about $61\%$ of
CoTA-DT's improvement over the reproduced T2DA-T despite training
its decision backbone from scratch. It also comes close to the full
model on Point-Robot ($-8.4\pm0.5$ versus $-7.9\pm0.4$).
In contrast, the much larger random-init 1B + LoRA variant reaches
only $916.8\pm36.0$ on Meta-World. One possible explanation is that
CLIP provides a useful task signal to the trainable small DT, whereas
LoRA must adapt frozen random features for both language and decision
modeling in the 1B control.

\section{Few-Shot Adaptation Modes}
\label{app:fewshot}

The few-shot results in Table~\ref{tab:fewshot_mixed} of the main
paper use our default mode (\emph{finetune} with $k{=}5$), where
$k$ denotes the number of target-task support trajectories.
We compare four modes that progressively add rollout-time latent
updates, language-grounded inversion, and policy fine-tuning:
\begin{itemize}[nosep, leftmargin=*]
  \item \textbf{static}: initialise $z=\bar z$ from the support
  trajectories and keep it fixed throughout the rollout;
  \item \textbf{online}: initialise $z=\bar z$ and refresh it from
  the agent's rollout history after collecting $H$ transitions;
  \item \textbf{inversion only}: refine $\bar z$ into $z_0$ via
  CLIP-space optimisation (Eq.~\ref{eq:clip_optimize}), then use
  the same rollout-time updates as \emph{online}, without updating
  the policy parameters;
  \item \textbf{finetune} (default): use the same inversion and
  rollout-time updates as \emph{inversion only}, and additionally
  fine-tune $\theta_{\mathrm{L}}$ and $f_{\mathrm{act}}$ for 100 Adam
  steps on the support set with $z_0$ held fixed.
\end{itemize}

\begin{table}[h]
\centering
\caption{Few-shot return with different adaptation modes on Point-Robot
and Meta-World (\emph{Mixed}, mean over 5 seeds). Best per column in
\textbf{bold}.}
\label{tab:fewshot_modes}
\setlength{\tabcolsep}{6pt}
\footnotesize
\begin{tabular}{l ccc c ccc}
\toprule
& \multicolumn{3}{c}{\textbf{Point-Robot} $\uparrow$}
&& \multicolumn{3}{c}{\textbf{Meta-World} $\uparrow$} \\
\cmidrule(lr){2-4} \cmidrule(lr){6-8}
Mode & $k{=}3$ & $k{=}5$ & $k{=}10$
&& $k{=}3$ & $k{=}5$ & $k{=}10$ \\
\midrule
static
& $-9.2$ & $-8.6$ & $-8.9$
&& $1281.3$ & $1295.1$ & $1290.5$ \\
online
& $-9.2$ & $-8.6$ & $-8.9$
&& $1295.8$ & $1318.0$ & $1325.2$ \\
inversion only
& $-8.4$ & $\mathbf{-6.9}$ & $\mathbf{-7.2}$
&& $1349.1$ & $1384.7$ & $\mathbf{1396.9}$ \\
finetune
& $\mathbf{-8.1}$ & $\mathbf{-6.9}$ & $-7.3$
&& $\mathbf{1352.6}$ & $\mathbf{1402.8}$ & $1395.3$ \\
\bottomrule
\end{tabular}
\end{table}

Three observations:
\emph{(i)} \emph{Static} and \emph{online} coincide on Point-Robot,
whose episode length equals the context horizon ($T=H=20$), leaving
no subsequent action to condition on a refreshed latent. On the
longer-horizon Meta-World tasks, online updates improve return over
\emph{static} at every tested $k$, suggesting that newly collected
transitions provide useful task information during rollout.
\emph{(ii)} Adding language-grounded inversion improves on
\emph{online} across both environments and all three support sizes.
At $k{=}5$, it raises Meta-World return from $1318.0$ to $1384.7$
without updating the policy parameters, highlighting the value of
a task-description-aligned initial latent.
\emph{(iii)} Policy fine-tuning adds a further $18.1$ return on
Meta-World at $k{=}5$, but leaves the Point-Robot mean unchanged;
at $k{=}10$, \emph{inversion only} slightly exceeds \emph{finetune}
on both environments. Fine-tuning is therefore an optional source
of improvement rather than a prerequisite for strong few-shot
results. At the default $k{=}5$, \emph{inversion only} already
reaches $-6.9$ on Point-Robot and $1384.7$ on Meta-World, above
the corresponding T2DA-T few-shot returns in
Table~\ref{tab:fewshot_mixed}.

\section{Visualisation Analysis}
\label{app:viz}

Beyond the qualitative t-SNE visualisation in Fig.~\ref{fig:tsne} of the main paper, we quantify the alignment by computing the rank-1 retrieval accuracy of the cross-modal nearest neighbour ($\bar z_i \leftrightarrow w_i$) on Point-Robot. The accuracy reaches $\mathbf{90\%}$ on \emph{Expert}, $\mathbf{88\%}$ on \emph{Medium}, and drops to $82\%$ on the heterogeneous \emph{Mixed} dataset. We attribute the Mixed-dataset drop to the bimodal trajectory distribution (medium $\cup$ expert) producing a less concentrated $\bar z_i$, which is itself an empirical finding about how dataset heterogeneity affects representation tightness. The same trend---highest retrieval accuracy on Expert, lowest on Mixed---also holds on Meta-World ($85\%$ / $81\%$ / $78\%$), confirming the observation is not Point-Robot-specific.

\section{Algorithm Summary}
\label{app:algo}

We present four algorithms that together describe CoTA-DT: \emph{(i)} multi-perspective representation learning (Algorithm~\ref{alg:repr}); \emph{(ii)} task-aware policy learning (Algorithm~\ref{alg:policy}); \emph{(iii)} zero-shot deployment via language-grounded inversion (Algorithm~\ref{alg:zeroshot}); and \emph{(iv)} few-shot adaptation via data-grounded initialisation and lightweight fine-tuning (Algorithm~\ref{alg:fewshot}).

\begin{algorithm}[!htbp]
\caption{CoTA-DT --- Multi-perspective representation learning}
\label{alg:repr}
\begin{algorithmic}[1]
\REQUIRE Per-task offline data $\{\mathcal{D}_i\}_{i=1}^{N}$; task descriptions $\{h_i\}_{i=1}^{N}$; frozen LLM backbone $\theta_{\mathrm{base}}$
\REQUIRE Loss weights $(\lambda_1, \lambda_2, \lambda_3, \lambda_4, \alpha)$; warm-up epochs $N_{\mathrm{warm}}$; sub-trajectory horizon $H$; latent dim $d_z$; alignment dim $d_a$
\STATE Initialise context encoder $E_\phi$, LoRA adapter $\theta_{\mathrm{L}}$, prediction heads $f_{\mathrm{rew}}, f_{\mathrm{dyn}}, f_Q$, continuous-token projectors $g_z, g_s, g_a, g_R$, alignment heads $\mathrm{Proj}_z, \mathrm{Proj}_h$, and learnable temperature $\kappa_\beta$
\FOR{epoch $=1, \dots, N$}
    \FOR{each minibatch of tasks $\mathcal{I}$}
        \STATE \textit{// (1) Trajectory branch}
        \FOR{each task $i \in \mathcal{I}$}
            \STATE Sample anchor sub-trajectory $\tau_{t-H:t}$ from $\mathcal{D}_i$; encode $z_t^{(i)} = E_\phi(\tau_{t-H:t})$
            \STATE Sample positive $\tau^{+}$ (nearby segment); encode $z_t^{+,(i)} = E_\phi(\tau^{+})$
            \STATE Sample $m$ sub-trajectories from $\mathcal{D}_i$; compute task-mean $\bar z_i = \frac{1}{m}\sum_{j=1}^{m} E_\phi(\tau_j^i)$
        \ENDFOR
        \STATE Treat in-batch latents from other tasks as negatives $\{z_k^{-}\}$ for InfoNCE
        \STATE \textit{// (2) Language branch}
        \STATE For each $i \in \mathcal{I}$: $v_i = \mathrm{LLM}_{\theta_{\mathrm{base}}, \theta_{\mathrm{L}}}(h_i)$
        \STATE \textit{// (3) Alignment}
        \STATE $u_i = \mathrm{Proj}_z(\bar z_i) / \|\mathrm{Proj}_z(\bar z_i)\|_2$, $\quad w_i = \mathrm{Proj}_h(v_i) / \|\mathrm{Proj}_h(v_i)\|_2$
        \STATE Compute similarity matrix $S_{ij} = u_i^\top w_j / \kappa_\beta$ for $i, j \in \mathcal{I}$
        \STATE \textit{// (4) Task-conditioned world model and Q learning}
        \STATE For each transition $(s_t, a_t, r_t, s_{t+1})$, predict $\hat r_t, \hat s_{t+1}, \hat R_t$ via shared LLM $+ \theta_{\mathrm{L}}$ on tokens $[g_z(z_t^{(i)}); g_s(s_t); g_a(a_t)]$
        \STATE \textit{// (5) Compute losses}
        \STATE $\mathcal{L}_{\mathrm{NCE}} \leftarrow$ InfoNCE over $(z_t^{(i)}, z_t^{+,(i)}, \{z_k^{-}\})$ with temperature $\kappa_\alpha$
        \STATE $\mathcal{L}_{\mathrm{align}} \leftarrow$ symmetric cross-entropy over $S_{ij}$ (Eq.~\ref{eq:clip_task_level})
        \STATE $\mathcal{L}_{\mathrm{rew}} \leftarrow \mathbb{E}[\|\hat r_t - r_t\|_2^2]$, $\quad \mathcal{L}_{\mathrm{dyn}} \leftarrow \mathbb{E}[\|\hat s_{t+1} - s_{t+1}\|_2^2]$, $\quad \mathcal{L}_{Q} \leftarrow \mathbb{E}[\|\hat R_t - R_t\|_2^2]$
        \STATE $\mathcal{L} \leftarrow \lambda_1 \mathcal{L}_{\mathrm{NCE}} + \mathbb{1}[\text{epoch} > N_{\mathrm{warm}}]\,\lambda_2 \mathcal{L}_{\mathrm{align}} + \lambda_3 (\mathcal{L}_{\mathrm{rew}} + \alpha \mathcal{L}_{\mathrm{dyn}}) + \lambda_4 \mathcal{L}_{Q}$
        \STATE Update $E_\phi, \theta_{\mathrm{L}}, \mathrm{Proj}_z, \mathrm{Proj}_h, g_*, f_{\mathrm{rew}}, f_{\mathrm{dyn}}, f_Q, \kappa_\beta$ via Adam
    \ENDFOR
\ENDFOR
\STATE \textbf{Output:} trained $E_\phi, \theta_{\mathrm{L}}, \mathrm{Proj}_z, \mathrm{Proj}_h, g_*$
\end{algorithmic}
\end{algorithm}

\begin{algorithm}[!htbp]
\caption{CoTA-DT --- Task-aware policy learning}
\label{alg:policy}
\begin{algorithmic}[1]
\REQUIRE Representation-learning checkpoint $\{E_\phi, \theta_{\mathrm{L}}, g_*\}$; per-task offline data $\{\mathcal{D}_i\}_{i=1}^{N}$; sub-trajectory horizon $H$; total steps $T_{\mathrm{train}}$
\STATE Initialise action head $f_{\mathrm{act}}$; load $\theta_{\mathrm{L}}, E_\phi, g_*$ from the representation checkpoint
\FOR{step $=1, \dots, T_{\mathrm{train}}$}
    \STATE Sample task $i \sim \mathcal{U}(1, N)$ and trajectory $\tau \sim \mathcal{D}_i$
    \STATE Sample a DT context window $(\hat R_{1:T_c}, s_{1:T_c}, a_{1:T_c})$ of length $T_c$ from $\tau$
    \STATE Compute task latent $z = E_\phi(\tau_{t-H:t})$ from the most recent $H$ steps (zero-padded if needed); held fixed within the window
    \STATE Forward DT with input $[z, g_R(\hat R_1), g_s(s_1), g_a(a_1), \dots]$ via LLM $+ \theta_{\mathrm{L}}$ and head $f_{\mathrm{act}}$
    \STATE $\mathcal{L}_{\mathrm{policy}} \leftarrow$ negative log-likelihood of predicted vs.\ ground-truth actions (Eq.~\ref{eq:policy_loss})
    \STATE Update $\theta_{\mathrm{L}}, f_{\mathrm{act}}, g_*$ via AdamW with gradient clipping
\ENDFOR
\STATE \textbf{Output:} adapted $\theta_{\mathrm{L}}, f_{\mathrm{act}}, g_*$
\end{algorithmic}
\end{algorithm}

\begin{algorithm}[!htbp]
\caption{CoTA-DT --- Zero-shot adaptation}
\label{alg:zeroshot}
\begin{algorithmic}[1]
\REQUIRE Trained model $\{E_\phi, \theta_{\mathrm{L}}, \mathrm{Proj}_z, \mathrm{Proj}_h, f_{\mathrm{act}}, g_*\}$; target task description $h_\star$; target return $\hat R_0$; episode length $T$; sub-trajectory horizon $H$; inversion steps $K_{\mathrm{inv}}$, inversion lr $\eta$
\STATE \textit{// Language-grounded latent initialisation}
\STATE Encode description: $v_\star = \mathrm{LLM}_{\theta_{\mathrm{base}}, \theta_{\mathrm{L}}}(h_\star)$
\STATE Project and normalise: $w_\star = \mathrm{Proj}_h(v_\star) / \|\mathrm{Proj}_h(v_\star)\|_2$
\STATE Initialise $z \leftarrow \mathbf{0} \in \mathbb{R}^{d_z}$
\FOR{$j = 1, \dots, K_{\mathrm{inv}}$}
    \STATE $u \leftarrow \mathrm{Proj}_z(z) / \|\mathrm{Proj}_z(z)\|_2$
    \STATE $\mathcal{L}_{\mathrm{inv}} \leftarrow 1 - u^\top w_\star$
    \STATE Update $z$ via one Adam step on $\nabla_z \mathcal{L}_{\mathrm{inv}}$ with learning rate $\eta$
\ENDFOR
\STATE Set $z_0 \leftarrow z$
\STATE \textit{// Self-adaptive rollout}
\STATE Receive initial state $s_1$; initialise rollout history $\mathcal{H} \leftarrow \emptyset$
\FOR{$t = 1, \dots, T$}
    \IF{$t \le H$}
        \STATE $z_t \leftarrow z_0$ \COMMENT{insufficient rollout history, use language-grounded latent}
    \ELSE
        \STATE $z_t \leftarrow E_\phi(\tau_{t-H:t-1})$ from $\mathcal{H}$ \COMMENT{re-encode from rollout}
    \ENDIF
    \STATE Predict action: $a_t \sim \pi_\theta(\cdot \mid \hat R_{\le t}, s_{\le t}, a_{<t};\, z_t)$
    \STATE Execute $a_t$, observe $r_t, s_{t+1}$; update RTG $\hat R_{t+1} \leftarrow \hat R_t - r_t$; append $(s_t, a_t, r_t)$ to $\mathcal{H}$
\ENDFOR
\end{algorithmic}
\end{algorithm}

\begin{algorithm}[!htbp]
\caption{CoTA-DT --- Few-shot adaptation}
\label{alg:fewshot}
\begin{algorithmic}[1]
\REQUIRE Trained model $\{E_\phi, \theta_{\mathrm{L}}, \mathrm{Proj}_z, \mathrm{Proj}_h, f_{\mathrm{act}}, g_*\}$; target task description $h_\star$; support set $\mathcal{D}_\star$ of size $k$; target return $\hat R_0$; inversion steps $K_{\mathrm{inv}}$, inversion lr $\eta$; fine-tune steps $K_{\mathrm{ft}}$, fine-tune lr $\eta_{\mathrm{ft}}$
\STATE \textit{// (1) Data-grounded latent initialisation}
\STATE Sample sub-trajectories $\mathcal{C} = \{\tau_1, \dots, \tau_{|\mathcal{C}|}\}$ from $\mathcal{D}_\star$
\STATE Compute aggregated latent: $\bar z = \frac{1}{|\mathcal{C}|} \sum_{\tau \in \mathcal{C}} E_\phi(\tau)$
\STATE Encode description: $v_\star = \mathrm{LLM}_{\theta_{\mathrm{base}}, \theta_{\mathrm{L}}}(h_\star)$, $\quad w_\star = \mathrm{Proj}_h(v_\star) / \|\mathrm{Proj}_h(v_\star)\|_2$
\STATE Initialise $z \leftarrow \bar z$ \COMMENT{data-grounded init in place of zero init}
\FOR{$j = 1, \dots, K_{\mathrm{inv}}$}
    \STATE $u \leftarrow \mathrm{Proj}_z(z) / \|\mathrm{Proj}_z(z)\|_2$
    \STATE Update $z$ via one Adam step on $\nabla_z (1 - u^\top w_\star)$ with learning rate $\eta$
\ENDFOR
\STATE Set $z_0 \leftarrow z$
\STATE \textit{// (2) Light-weight policy fine-tune}
\FOR{$j = 1, \dots, K_{\mathrm{ft}}$}
    \STATE Sample DT window $(\hat R_{1:T_c}, s_{1:T_c}, a_{1:T_c})$ from $\mathcal{D}_\star$
    \STATE Forward policy with $z = z_0$ held fixed; compute $\mathcal{L}_{\mathrm{policy}}$ (Eq.~\ref{eq:policy_loss})
    \STATE Update $\theta_{\mathrm{L}}, f_{\mathrm{act}}$ via Adam at lr $\eta_{\mathrm{ft}}$ (all other parameters frozen)
\ENDFOR
\STATE \textit{// (3) Self-adaptive rollout}
\STATE Run the self-adaptive rollout phase of Algorithm~\ref{alg:zeroshot} using the fine-tuned policy and the data-grounded $z_0$
\end{algorithmic}
\end{algorithm}

\newpage

\end{document}